\documentclass[11pt,a4paper]{article}

\usepackage[T1]{fontenc}
\usepackage[utf8]{inputenc}
\usepackage[a4paper,margin=2.5cm]{geometry}
\usepackage{amsmath,amssymb,amsfonts}
\usepackage{graphicx}
\usepackage{booktabs}
\usepackage{multirow}
\usepackage{array}
\usepackage{caption}
\usepackage{subcaption}
\usepackage{float}
\usepackage[round,sort&compress]{natbib}
\usepackage{xcolor}
\usepackage[colorlinks=true,linkcolor=blue,citecolor=blue,urlcolor=magenta]{hyperref}

\title{Agentic System as Compressor:\\ Quantifying System Intelligence in Bits}
\author{%
  Zihan Qin\thanks{Equal contribution.} \\
  Peking University
  \and
  Hongrui Zhang\footnotemark[1] \\
  Peking University
}
\date{}

\begin{document}
\maketitle

\begin{abstract}
Large language models are turning from isolated predictors into \textbf{agentic systems}: they call tools, retrieve evidence, obey environment constraints, use verifiers, and complete tasks through search and multi-turn interaction. We adopts an analytical viewpoint based on ``compression is intelligence'': under a fixed task distribution, interface, and compute budget, a stronger agentic system lets a target object be reconstructed with \textbf{fewer bits}. We operationalize the measure with \textbf{arithmetic coding}, \textbf{seed coding}, and a \textbf{fallback}, and evaluate it in five settings: reversed text, chess moves, protein sequences, retrieval-augmented question answering, and semantic story compression; in all of them agentic components reduce codelength. These small, controlled experiments cover component types typical of real agentic systems, show that codelength can analyze how components, observers, and budgets change residual uncertainty, and offer guidance for evaluating real agent systems.
\end{abstract}

\section{Introduction}

Large language models appear less and less as isolated predictors and more and more embedded in \textbf{agentic systems} that can call tools, retrieve evidence, read feedback, and interact with an environment \citep{yao2023react,schick2023toolformer,lewis2020rag,shinn2023reflexion,yang2024sweagent}. In real deployments, intelligent behavior often arises from the whole workflow. That is, the capability deployed and used today is no longer just ``model intelligence'' but \textbf{system intelligence} formed jointly by the model, tools, environment interfaces, and interaction process.

This creates a tension for evaluation. Standard benchmarks usually report accuracy, success rate, pass rate, or an aggregate score \citep{liang2022helm,srivastava2022bigbench}; these metrics indicate whether the system finally completes the task, but say little about where the capability comes from. When a system grows stronger, is it because the model itself predicts better, or because a retriever supplied evidence, a rule-based environment ruled out illegal actions, a verifier filtered out bad trajectories, or a search process found better candidates? How much does a given component actually contribute? When a task is hard, the success rate can stay near zero for a long time; both systems then ``almost fail,'' yet their reductions of the residual uncertainty over the target space may already differ enormously.

The compression-as-intelligence viewpoint offers a finer analytical framework \citep{cover2006elements,witten1987arithmetic,deletang2023language,huang2024compression}. Classical information theory shows that a predictive distribution can be turned into a lossless compressor and vice versa \citep{cover2006elements,witten1987arithmetic}; a language model induces token-level codelengths, making log-loss and bits-per-byte an operational compression rate \citep{deletang2023language}; and cross-model results further show that the compression rate is nearly linearly correlated with downstream capability \citep{huang2024compression}. These works, however, mainly measure the bare model itself, without bringing tools, environment, retrieval, and the interaction process into the object of measurement.

This paper therefore adopts \textbf{agentic codelength} as a system-level working measure. Intuitively, once the task distribution, observation standard, system interface, and coding protocol are fixed, the encoder need not transmit the target object in full; it sends only a per-instance code that lets a decoder sharing the same system conditions reconstruct the target. The more intelligent a system, the fewer residual bits it should require to reconstruct objects from the same distribution. Bare-model compression is just a special case: a system can choose to call no tools, do no retrieval, and no search, degenerating into ordinary model compression; conversely, when environment interaction provides usable structure, the system can turn that structure into a shorter per-instance code.

To estimate this system compression ability, we combine the following three schemes into an executable compress--decompress protocol. First, \textbf{arithmetic coding} accumulates the negative log-likelihood (NLL) on a given output, turning model predictions into codelength, for cases where a specific output must be encoded exactly. Second, \textbf{seed coding} targets tasks in deterministic or replayable environments where qualifying outputs are dense enough within the sampling budget: it transmits a latent description plus the rank of the first successful random seed. Third, when seed coding fails within a finite budget, it \textbf{falls back} to arithmetic coding, so that every sample can be encoded successfully.

We validate the framework in five settings: reversed text, chess, protein sequences, retrieval-augmented question answering, and semantic story compression. These settings respectively isolate several types of information source typical of agentic systems: deterministic tools, rule-based environments, scientific priors, verifier feedback, retrieval side information, semantic observers, and the sampling budget. Reversed text provides a minimal sanity check, testing whether a shared transform can move per-instance structure into the decoder; chess tests how environment constraints shrink the space of legal candidates; the protein experiments test both whether codelength can read out the value of template priors when the success rate has almost no resolution, and whether verifier feedback can reduce blind-search cost in functionally equivalent generation; RAG and story compression respectively test how external-evidence relevance, the observation standard, and the compute budget change the residual codelength. In short, the goal of this set of experiments is not to show that some agent achieves the highest score on a single task, but to verify that agentic codelength can serve as a unified accounting tool for residual uncertainty across different information sources and different success criteria.

\begin{figure}[htbp]
  \centering
  \includegraphics[width=\linewidth]{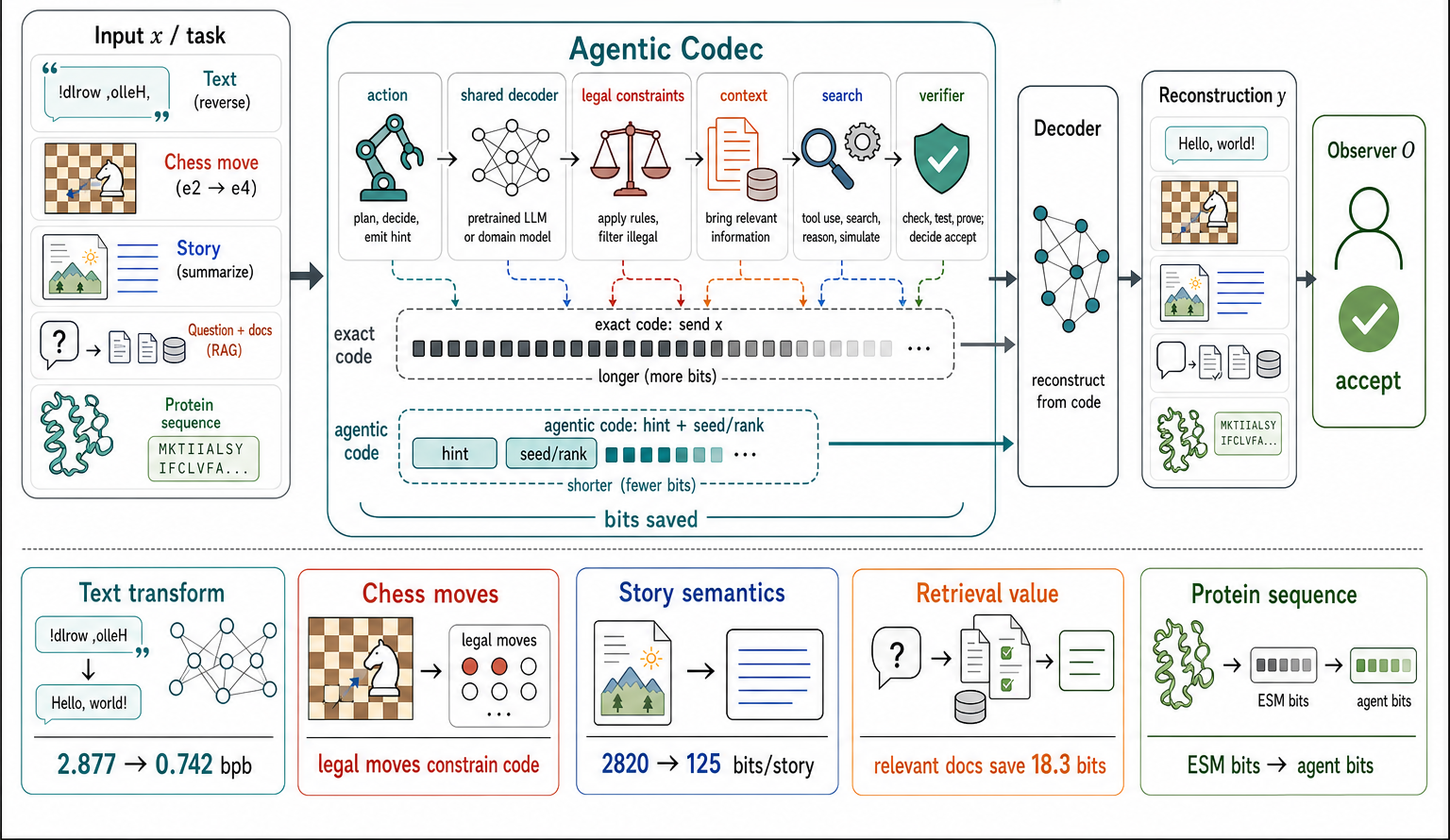}
  \caption{Overview of agentic compression. Tools, constraints, context, search, and verifiers act as shared decoding resources, reducing the per-instance hint, seed/rank, and residual bits needed for a reconstruction accepted by the observer.}
  \label{fig:overview0}
\end{figure}

Our contributions are as follows. First, we are the first to understand the ``compression is intelligence'' view from the perspective of \textbf{agentic systems} rather than isolated models. Second, we give a \textbf{marginal bit value} metric for components, quantitatively analyzing the contribution of a specific component in a system via the average codelength reduction. Third, we combine arithmetic coding, seed replay, and fallback coding into an executable compress--decompress protocol. Fourth, we validate the effectiveness of this metric for analyzing system intelligence across five classes of tasks.

\section{Related Work}

\subsection{Compression, intelligence, and ``language models as compressors''}

``Compression measures understanding'' is a classical idea: in an appropriate sense, prediction and compression are interconvertible, and the two are essentially equivalent \citep{cover2006elements,witten1987arithmetic}. Recent discussions of large models, including Sutskever's talk \emph{An Observation on Generalization}, popularized a sharper operational reading: a predictor that assigns higher probability to a data distribution also induces a shorter code for samples from that distribution \citep{sutskever2023observation}.

This idea has a more systematic formalization in MDL (minimum description length) and universal coding. The Kraft--McMillan inequality shows that a uniquely decodable codelength function corresponds to a probability budget; and the log-loss risk $-\log q(x)$ of a probabilistic model $q$ is precisely the ideal description length it induces. Hence, learning within a model class $\mathcal{M}=\{q_\theta\}$ is not only a search for parameters with low predictive risk, but also a search for a description language that assigns short codes to data. The MDL principle writes this relationship as a two-part code or universal-coding criterion: choose an explanation by the sum of the model description length $L(\theta)$ and the residual data description length $L(x\mid\theta)\simeq-\log q_\theta(x)$ \citep{rissanen1978modeling,barron1998mdl,grunwald2007mdl}. From this perspective, generalization can be understood as structure discovery in the compression sense: reusable regularities enter the shared model description, while sample-specific content remains in the residual codelength.

Our most direct predecessor is the ``language modeling is compression'' line of work: an autoregressive model induces token-level codelengths, so log-loss and bits-per-byte provide an operational compression rate \citep{deletang2023language}. Cross-model measurements further show that the compression rate is nearly \textbf{linearly} correlated with downstream capability---the shorter the code, the stronger the model \citep{huang2024compression}---suggesting that compression ability is a useful measure of intelligence. These works, however, measure the compression rate of the \textbf{bare model} itself. They do not touch one reality: today's models often do not work in isolation---they are embedded in agentic systems and, through tools, feedback, and multi-turn environment interaction, achieve capabilities well beyond the model proper. This paper aims precisely to extend the compression viewpoint from isolated models to entire agentic systems.

\subsection{Agentic systems, tools, retrieval, and feedback}

Modern work on agentic systems studies how models call tools, retrieve information, receive feedback, repair their own outputs, and act through multi-turn interaction. ReAct interleaves reasoning and acting \citep{yao2023react}; Toolformer teaches a model to call APIs \citep{schick2023toolformer}; RAG conditions generation on retrieved documents \citep{lewis2020rag}; Reflexion uses verbal feedback to improve subsequent attempts \citep{shinn2023reflexion}. On the reasoning side, chain-of-thought \citep{wei2022cot}, self-consistency \citep{wang2023selfconsistency}, and tree-of-thoughts \citep{yao2023tot} search over intermediate states.

Recently, these techniques have been consolidated into a deliberate engineering practice---\textbf{harness engineering}: beyond strengthening the model proper, it emphasizes polishing the surrounding tools, environment interfaces, feedback loops, and context management, and improving the environment rather than blaming the model when an agent fails \citep{openai2026harness}. The OpenAI team, using the coding agent Codex as the sole executor and continually polishing the harness, produced production code on the order of a million lines within a few months \citep{openai2026harness}; on the open-source side, runtimes such as OpenClaw and Hermes Agent package ``model + tools + environment + multi-turn loop'' into a directly deployable harness and quickly became among the most watched projects in the community \citep{openclaw2026,hermes2026}.

\subsection{Observational equivalence, semantic compression, and perceptual compression}

Many compression tasks do not require byte-exact recovery, only that the reconstruction preserve key properties in the eyes of the target observer: perceptual codecs such as JPEG and MP3/AAC exploit the limited sensitivity of human vision and hearing to discard imperceptible detail \citep{wallace1991jpeg,brandenburg1999mp3}, and perceptual-quality studies likewise point out a trade-off between pointwise distortion and ``looking natural'' \citep{blau2018perception}. Semantic compression extends this idea to structured content: the reconstruction need not be identical token-by-token or byte-by-byte, only preserve the meaning, entities, events, function, or answer that the task cares about; relevant precedents include semantic compression in databases \citep{jagadish1999semantic} and prompt/context compression in the LLM setting \citep{jiang2023llmlingua}. This paper further expresses such ideas as compression under \textbf{observational equivalence}; see Section~\ref{sec:obseq}.

\section{Theory}

We proceed in four steps. We first define agentic systems (Section~\ref{sec:agentdef}), then give conditional compression (Section~\ref{sec:condcomp}) and observational equivalence (Section~\ref{sec:obseq}). We then discuss \textbf{agentic coding} and component value within a fixed actual system (Section~\ref{sec:agentcoding}).

\subsection{Definition of an agentic system}\label{sec:agentdef}

We model an agentic system as a triple
\[
\mathcal{A}=(P,\,M,\,E),
\]
where $P$ is a program containing the model interface and the environment interface; $M$ is one or more \textbf{model interfaces} that map text to token distributions; and $E$ is a set of \textbf{environment interfaces}, including retrievers, compilers, external APIs, and other tool-call interfaces. A resource constraint $C$ is given as an external budget index at run and evaluation time, which can limit the total number of execution steps, model calls, tool calls, context length, wall-clock time, memory, cost, or any other constraint we care about. During execution, $P$ may query $M$ and $E$, update its state according to their return values, and continue computing until the resource constraint is exhausted. Hereafter we write $\mathcal{A}_C$ for the run of system $\mathcal{A}$ under budget $C$.

We stress that, in our setting, the \textbf{encoder} system $\mathcal{A}_e$ and the \textbf{decoder} system $\mathcal{A}_d$ are in general \textbf{asymmetric}. This differs from ordinary language-model arithmetic coding, in which a single model is bound symmetrically to the encoder and decoder, and coding and decoding are essentially tied to one and the same model distribution.

\subsection{Conditional compression}\label{sec:condcomp}

Many of the tasks we discuss are naturally modeled as conditional compression tasks: an object does not appear in isolation but is given together with a problem statement, task specification, or other public context. Formally, data are given as condition--target pairs $(c,x)\sim\mathcal{D}$, where $c$ is \textbf{distribution-dependent side information} shared and fixed by the encoder and decoder before any object is encoded. In our tasks, $c$ is usually the problem statement or task specification. \textbf{Conditional compression} measures codelength given $c$: the encoder sends a per-instance description $d=\mathrm{Enc}(x;c)$, and the decoder computes $\mathrm{Dec}(d;c)$ and must recover $x$ in the observational sense defined in Section~\ref{sec:obseq}. Only $d$ counts toward codelength; $c$ does not. We therefore measure the conditional codelength $L(x\mid c)$, not the marginal codelength $L(x)$. This cleanly separates the background shared by both sides from the extracted information that must be supplied per instance. It follows the conditional setting of Hsiao--Lu--Reyzin, in which the compressor and decompressor have access to side information \citep{hsiao2007conditional}, and makes the codelengths of different systems comparable under the same $c$.

\subsection{Observational equivalence}\label{sec:obseq}

Let $\mathcal{O}$ be a family of observations, where each $o\in\mathcal{O}$ maps an artifact to an observed value. We call $y$ and $x$ \textbf{observationally equivalent} when every observation agrees:
\[
y\approx_{\mathcal{O}} x \;\Longleftrightarrow\; \forall\, o\in\mathcal{O},\; o(y)=o(x).
\]
Reconstruction succeeds as long as the decoder outputs any $y$ satisfying $y\approx_{\mathcal{O}}x$. We write the equivalence class as $[x]_{\mathcal{O}}=\{y:y\approx_{\mathcal{O}} x\}$. Exact recovery is the special case where $\mathcal{O}$ distinguishes all points, in which the condition reduces to $y=x$. When $\mathcal{O}$ preserves task-relevant observations rather than bytes, the same formalization becomes a form of semantic compression.

For example, the protein task (Section~\ref{sec:claim1} and Appendix~\ref{app:protein}) cares only about \textbf{functional equivalence}. A reconstruction $y$ is accepted if and only if it agrees with the original sequence $x$ (i.e.\ $o(y)=o(x)$) on the boolean observations specified by the verifier; residue-by-residue identity is not required.

\paragraph{Verifiers as a special case.} A verifier that checks whether a reconstruction $y$ agrees with $x$ on properties $\{o_i\}$ is exactly the observational equivalence $y\approx_{\mathcal{O}}x$ with $\mathcal{O}=\{o_i\}$, whose acceptance set is the class $[x]_{\mathcal{O}}$. An ``absolute'' verifier $V(y)\in\{0,1\}$ independent of $x$ also fits: since the original sequence satisfies $V(x)=1$, writing $o(y)=V(y)$ makes the acceptance condition $V(y)=1$ equivalent to $o(y)=o(x)$. Hence the codelength definition in Section~\ref{sec:agentcoding} need not treat verifiers as a separate case.

\subsection{Agentic coding and component value}\label{sec:agentcoding}

We consider a fixed coding protocol in which the encoder and decoder are two agentic systems,
\[
\mathcal{A}_e=(P_e,M_e,E_e),
\qquad
\mathcal{A}_d=(P_d,M_d,E_d).
\]
Here $\mathcal{A}_e$ is the encoding-side system and $\mathcal{A}_d$ the decoding-side system; the two may use different programs and interfaces and are constrained by budgets $C_e$ and $C_d$ respectively. The encoder is usually allowed stronger offline search, such as enumerating candidate latents or random seeds; the decoder only replays the protocol given the message, the public condition, and its own budget. To make decoding verifiable, the protocol specifies the public condition $c$ visible to both ends, the observation standard $\mathcal{O}$, the model/tool versions, the randomness convention, and the interaction interfaces. We write the whole protocol as
\[
\Pi=(\mathcal{A}_e,\mathcal{A}_d,C_e,C_d,\mathcal{O}).
\]

In the conditional-compression setting, the encoding-side program reads the object $x$ to be encoded and the public condition $c$, and produces a message
\[
m=P_e(x;c,M_e,E_e,C_e).
\]
The decoding-side program receives only $m$ and the public condition $c$, then calls its own model interface, environment interface, and budget to output
\[
\hat{x}=P_d(m;c,M_d,E_d,C_d).
\]
If $\hat{x}\approx_{\mathcal{O}}x$, the message successfully reconstructs $x$ under the observation standard $\mathcal{O}$. Under the fixed protocol, we write the operational codelength of this instance as
\[
L_{\Pi}(x\mid c)=|m|.
\]
For a data distribution $\mathcal{D}$, the system's average codelength is
\[
\bar L_{\Pi}(\mathcal{D})
=
\mathbb{E}_{(c,x)\sim\mathcal{D}}
L_{\Pi}(x\mid c).
\]
The average codelength measures this fixed protocol's compression ability on the distribution: the shorter the average codelength, the less residual information each instance still needs after the public condition, decoding-side system, observation standard, and budget have all been given. For arithmetic coding realized by a predictive distribution, this quantity also equals the model's average negative log-likelihood under the protocol; it can therefore be interpreted as a codelength characterization of predictive ability.

To measure the value of a component $g$, we compare the average codelength of the same base protocol with and without $g$. Let $\Pi_g$ denote the protocol containing component $g$, and $\Pi_{\setminus g}$ the baseline protocol with that component removed and the remaining data, observer, model version, and budget kept as fixed as possible; then we define
\[
\Delta_g
=
\mathbb{E}_{(c,x)\sim\mathcal{D}}\,L_{\Pi_{\setminus g}}(x\mid c)
-
\mathbb{E}_{(c,x)\sim\mathcal{D}}\,L_{\Pi_g}(x\mid c).
\]
$\Delta_g$ is a marginal quantity: it measures the average codelength reduction from adding component $g$ to a given baseline protocol, and can thus measure that component's contribution to the system's prediction and reconstruction ability. Along a fixed order of component additions, the successive marginal values sum to the total reduction; but when components interact, an individual $\Delta_g$ depends on the baseline and the addition order.

\section{Coding Schemes}

The core problem in designing a coding scheme is that a model's output is random. When an agentic system runs as the decoder, how do we \textbf{pin down} the model's output each time with as short a code as possible? We introduce two methods---\textbf{arithmetic coding} and \textbf{seed coding}---for the cases ``a specific output must be encoded exactly'' and ``there are many qualifying outputs,'' respectively, and fall back to arithmetic coding when seed coding fails.

\subsection{Arithmetic coding}

Arithmetic coding is used to \textbf{exactly encode} a given piece of text: under context $c$, the model induces token-level distributions $P_M(\cdot\mid x_{<t},c)$, and arithmetic coding takes their conditional probabilities token-by-token on the reference $x$, transmitting $x$ unambiguously (we omit the details; see ``language modeling as compression'' \citep{deletang2023language}). With the standard prefix-unambiguous termination, its codelength is $L_{\mathrm{LM}}(x\mid c)+O(1)$, concretely $\lceil L_{\mathrm{LM}}(x\mid c)\rceil+1$, and thus exceeds the true NLL by fewer than $2$ bits (Appendix~\ref{app:ac}). Hereafter we \textbf{mainly report NLL}; this per-instance constant difference barely affects the conclusions.

\subsection{Seed coding}

If a desired output is dense enough under the LLM's output distribution---i.e.\ very likely to be hit within the given sampling budget---then \textbf{seed coding} is usually more bit-efficient than arithmetic coding.

The randomness of each model generation is determined by a random seed; given the seed, the output is deterministic. Let the sampling budget be $N$ seeds (numbered $0,1,\ldots,N-1$). The encoder tries them in order from $0$, takes a seed within budget whose output $y\approx_{\mathcal{O}}x$, and transmits its index with a \textbf{fixed-length $\left\lceil \log_2 N \right\rceil$-bit code}; the decoder replays the same seed and obtains the same qualifying output. Hence the coding cost of this instance is just this fixed-length index of $\left\lceil \log_2 N \right\rceil$ bits. Because the sampling budget is generally low, this codelength generally does not dominate.

The protocol premise of seed coding is \textbf{replayability}: the encoder and decoder must reproduce the same random trajectory under the same system conditions, including a fixed model, sampling algorithm, random seed, tool returns, and environment state. This paper uses seed coding only in deterministic or replayable environments; nondeterministic external services are out of scope. Seed coding is merely one way to operationalize an achievable codelength upper bound under a fixed protocol, and does not affect the theoretical analysis framework.

\subsection{Fallback}

If \textbf{no} seed within the sampling budget makes the output qualify, seed coding fails for that instance, and we fall back to arithmetic coding---which can always encode the reference $x$ exactly, giving a definite codelength $L_{\mathrm{fallback},i}$. The two paths are distinguished by a \textbf{1-bit selector}: the decoder reads this bit first, then decodes $x$ via seed coding or arithmetic coding accordingly. Thus the codelength of each instance is
\[
L_i=1+\min\{L_{\mathrm{agent},i},\,L_{\mathrm{fallback},i}\},
\]
where $L_{\mathrm{agent},i}=\left\lceil \log_2 N \right\rceil$ (hit within budget), and on a miss we set $L_{\mathrm{agent},i}=\infty$, with arithmetic coding as the backstop.

\section{Experiments}

We use five sets of experiments to argue the following three claims. These experiments cover component types typical of real agentic systems (tools, rule-based environments, scientific templates, retrievers, semantic observers, and the search budget), but are deliberately kept small and controlled, so as to isolate mechanisms and test whether the framework can analyze system components. Accordingly, our experiments should be read as mechanistic arguments and exploratory evidence, not as a large-scale benchmark or a final evaluation of a deployed system.

\begin{itemize}
  \item \textbf{Claim 1 (main claim: environment interaction improves compression, and the gain is quantifiable).} Providing the model with harness components such as tools, rule-based environments, scientific templates, verifier feedback, and retrievers, and allowing the system to interact with these environment resources, can improve the system's compression ability---fewer residual bits are needed to reconstruct the target; this improvement not only exists but can be quantitatively analyzed via the average bits saved. Jointly argued by \textbf{A reversed text}, \textbf{B chess}, \textbf{C/C$'$ protein}, and \textbf{D retrieval}.
  \item \textbf{Claim 2 (observation granularity affects compression).} The tightness of the observation standard changes the set of acceptable reconstructions, hence the system's compression ability, so the evaluation standard must be reported together with compression ability. Argued by the semantic-observer ladder experiment in the \textbf{E story} setting.
  \item \textbf{Claim 3 (a trade-off between compression and compute).} There is a trade-off between a system's compression ability and the compute invested. Demonstrated by the rollout--codelength Pareto frontier experiment in the \textbf{E story} setting.
\end{itemize}

Table~\ref{tab:claims} collects this mapping together with each component's bit value. \textbf{Naming convention: letters identify experiments (A--E); digits identify conditions within an experiment; in component-ablation experiments, \texttt{0} denotes the baseline without the component under test.} The full data, models, and protocols of each experiment are in the corresponding appendix subsection; the main text describes only design and argument.

\begin{table}[t]
\centering
\caption{Claims, experiments, and component values.}
\label{tab:claims}
\footnotesize
\setlength{\tabcolsep}{4pt}
\renewcommand{\arraystretch}{1.25}
\begin{tabular}{@{}p{2cm}p{2.5cm}p{3.0cm}p{4cm}p{2.1cm}c@{}}
\toprule
Claim & Exp. & Setup / condition change & Codelength change / trend & Component value & App. \\
\midrule
\multirow{5}{2.5cm}{\textbf{1}\,interaction improves compression, quantifiably}
 & \textbf{A} reversed text & A0 encode reversed string directly $\rightarrow$ A1 shared reverse transform & $2.877 \rightarrow 0.742$ bits/byte & $2.135$ bits/byte & \ref{app:reverse} \\
 & \textbf{B} chess & B0 plain SAN text $\rightarrow$ B1 legal-move environment & $9.828 \rightarrow 6.545$ bits/move & $3.283$ bits/move & \ref{app:chess} \\
 & \textbf{C} protein & C0 plain ESM2 $\rightarrow$ C2 calibrated template & $200.65 \rightarrow 116.80$ bits/seq & $83.85$ bits/seq & \ref{app:protein} \\
 & \textbf{C$'$} protein (functional eq.) & C$'$0 no-feedback sampling $\rightarrow$ C$'$1 verifier-feedback repair & $402.99 \rightarrow 27.58$ bits/seq & $375.41$ bits/seq & \ref{app:protein} \\
 & \textbf{D} retrieval (RAG) & D0 retrieval failure $\rightarrow$ D1 relevant docs & $24.46 \rightarrow 6.17$ bits/answer & relevant $18.29$; distractor $2.95$ & \ref{app:rag} \\
\midrule
\textbf{2}\, observation granularity affects compression & \textbf{E} story (semantic ladder) & observer tightened weak$\to$strong; E1/E2 replay the same candidate set & see Fig.~\ref{fig:ladder} & codelength rises monotonically as observer tightens & \ref{app:story} \\
\midrule
\textbf{3}\,compression vs.\ compute trade-off & \textbf{E} story (budget sweep) & summary rollout budget $T=1\rightarrow64$ & $292.1 \rightarrow 127.5$ bits/story & --- see \S\ref{sec:claim3} --- & \ref{app:story} \\
\bottomrule
\end{tabular}
\end{table}

Unless otherwise noted, arithmetic-coding codelengths are reported as NLL (differing from the actual codelength by only a constant, Appendix~\ref{app:ac}); the component value in each comparison follows the definition in Section~\ref{sec:agentcoding}, i.e.\ the average codelength reduction before and after adding the component. Lower codelength is always better.

\begin{figure}[htbp]
  \centering
  \includegraphics[width=\linewidth]{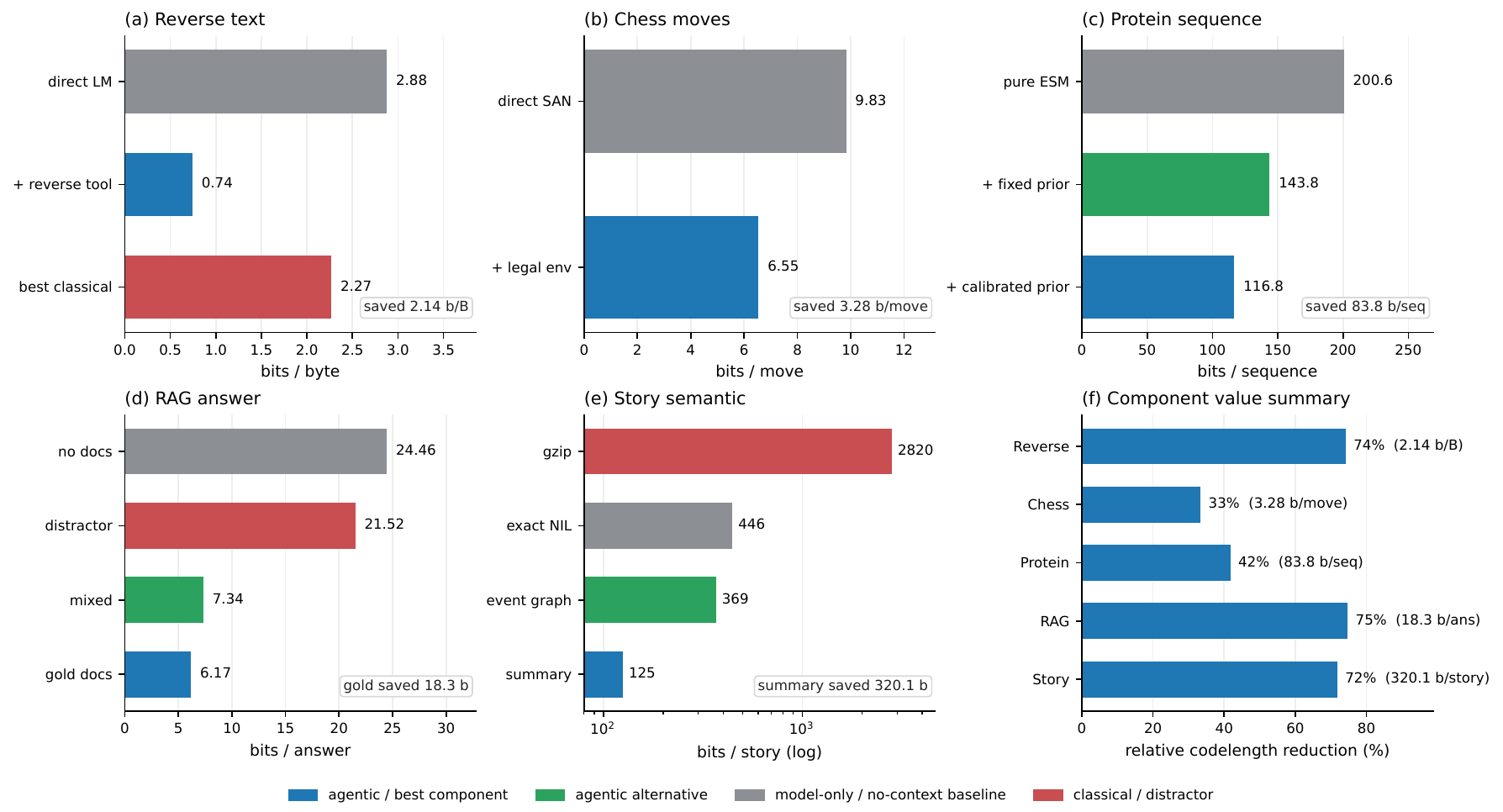}
  \caption{Bit value of agentic components across the five experiments (A--E). The first five panels show the absolute codelength before and after adding the component; the last panel summarizes the relative reduction.}
  \label{fig:bits}
\end{figure}

\subsection{Claim 1 (main claim): harness interaction improves compression, and the gain is quantifiable}\label{sec:claim1}

We argue Claim 1 with four controlled comparisons: A and B are minimal mechanism demonstrations, while C and D use real data to analyze the scientific-template and retrieval components, further showing that this codelength reduction stays quantifiable even when the success rate has almost no resolution (C) or when retrieval relevance varies continuously (D). Full experimental designs are in Appendices~\ref{app:reverse}--\ref{app:rag}.

\paragraph{A reversed text (deterministic tool).} The target distribution is the first $1\,\mathrm{MiB}$ of natural text from enwik8, processed by a deterministic line-wise reversal $R$, requiring exact byte recovery, with the model SmolLM2-360M. The baseline \textbf{A0} arithmetic-codes the reversed string directly, averaging $2.877$ bits/byte. \textbf{A1} lets the encoder and decoder share $R$: the encoder first uses $R$ to restore the reversed string to canonical text $u$, then arithmetic-codes it. The result: A1's codelength drops to $\mathbf{0.742}$ bits/byte, far below A0 and even below a general-purpose compressor such as gzip; the bit value of the reverse tool is $2.877-0.742=\mathbf{2.135}$ bits/byte---the deterministic tool reduced the system's residual codelength. This is the simplest demonstration of Claim 1.

\paragraph{B chess (rule-based environment).} The data are $100$ games, $6{,}005$ half-moves in total, deterministically sampled from the Lichess 2013-01 shard, with moves in standard algebraic notation (SAN). The baseline \textbf{B0} treats a whole game as ordinary text and arithmetic-codes it token-by-token, averaging $9.828$ bits/move. \textbf{B1} introduces a rule-based environment: at each position it calls \texttt{python-chess} to enumerate the legal-move set $L_t$, renormalizes the model probability over $L_t$ only to $Q(a)=P_M(a)/P_M(L_t)$, and takes the move's codelength as $-\log_2 Q$. The legal-move environment lowers the codelength to $\mathbf{6.545}$ bits/move, a bit value of $\mathbf{3.283}$ bits/move.

\paragraph{C protein (scientific template).} This experiment implements the protein task as a controlled \textbf{masked conditional reconstruction} agent, measuring how much a homologous template lowers the uncertainty of unobserved residues. On $100$ real protein sequences ($50$ each from PF00069, the protein kinase domain, and PF00096, the C2H2 zinc finger), the ESM2 masked language model, and leave-one-out template retrieval (full setup in Appendix~\ref{app:protein}), the three systems share the same target sequence, the same visible context, the same mask positions, and the same model, differing only in the conditional distribution $Q_i$ used at masked positions, so the codelength difference can be attributed to the homologous-template component itself. We add the ``homologous template'' component along a three-rung ladder, and the codelength shortens at each rung:

\begin{itemize}
  \item \textbf{C0 (plain ESM2, no template).} Masked positions are encoded directly with ESM2's distribution $p_{\text{ESM}}$ conditioned on the visible residues, averaging $200.65$ bits/seq.
  \item \textbf{C1 (fixed homologous template).} The system ranks candidate templates from the leave-one-out reference set by visible-site identity, visible 3-mer Jaccard, and a length penalty, locally aligns the best template to the masked target sequence, and forms a per-position residue prior $p_{\text{prior}}$ at masked positions. The coding distribution is $(1-\alpha)\,p_{\text{ESM}}+\alpha\,p_{\text{prior}}$, but the template reliability is fixed and $\alpha$ is capped at $0.50$. The codelength drops to $143.83$ bits/seq.
  \item \textbf{C2 (calibrated homologous template).} The system still uses the same template retrieval and alignment process, but $\alpha$ is no longer fixed; instead it estimates template-match reliability using visible residues only, and selects the mixing strength on a small subset of visible validation sites; masked positions are used only at final scoring. This rule can be recomputed by the decoding end from the shared visible context, so no extra transmission of template reliability is needed, and the cap on $\alpha$ is relaxed to $0.95$. The codelength further drops to $\mathbf{116.80}$ bits/seq.
\end{itemize}

Hence the total bit value of the calibrated homologous-template tool is $200.65-116.80=\mathbf{83.85}$ bits/seq, of which ``upgrading from fixed to calibrated'' (C1$\rightarrow$C2) alone contributes $27.03$ bits/seq---showing that a component's \textbf{reliability calibration} can itself be quantified in bits. Over $100$ sequences, C2 beats C0 with $71$ wins, $24$ losses, and $5$ ties.

\paragraph{C$'$ protein (functional equivalence and verifier feedback).} We further organize a functional-equivalence experiment. The data are still PF00069 and PF00096, $50$ each, $100$ real sequences in total, with leave-one-out to prevent target leakage. The observer requires candidate length in $80$--$300$ aa, satisfaction of family motif constraints, and that the top family hit of an HMMER scan be the target Pfam family with $e<10^{-5}$ and coverage $>0.6$. We compare two systems that differ only in the verifier-feedback component: \textbf{C$'$0 MODEL\_ONLY} samples templates from the same-family leave-one-out reference and generates candidates at mutation rate $0.30$, searching within $32$ ranks for the first accepted candidate and sending a seed code, but never reading the reason for failure; \textbf{C$'$1 AGENTIC} uses exactly the same initial proposals and random seeds, but reads the missing-motif information returned by the verifier only when a candidate fails the motif check, and performs at most $2$ rounds of motif repair before submitting to the same HMMER observer. This setup confines the gain to verifier feedback + repair, rather than changing the model, data, or sampling budget.

\begin{table}[t]
\centering
\caption{Shared-context codelength in functionally equivalent protein generation.}
\label{tab:protein-fe}
\begin{tabular}{lrrr}
\toprule
System & Success rate & Fallback rate & Mean conditional codelength \\
\midrule
C$'$0 \texttt{MODEL\_ONLY} & $68.0\%$ & $32.0\%$ & $402.99$ \\
C$'$1 \texttt{AGENTIC} & $\mathbf{98.0\%}$ & $\mathbf{2.0\%}$ & $\mathbf{27.58}$ \\
\bottomrule
\end{tabular}
\end{table}

The results show that verifier feedback significantly shortens the conditional codelength needed for functionally equivalent recovery. C$'$1 raises the success rate from $68.0\%$ to $98.0\%$ and lowers the mean conditional codelength from $402.99$ to $27.58$ bits/seq, saving $375.41$ bits/seq on average. In per-sample paired comparison, C$'$1 achieves $48$ wins, $0$ losses, and $52$ ties.

\paragraph{D retrieval-augmented generation (retriever).} The component being measured is the \textbf{retriever}; we implement four agentic systems D0--D3 that differ only in the retrieval component: D0 simulates a retrieval failure (no valid documents), D1 a retriever returning relevant documents, D2 a retriever returning distractor documents, and D3 a realistic scenario---a retriever returning a mix of true and distractor documents. We measure the bits the system needs to compress the answer, i.e.\ the conditional NLL $-\log_2 P(\text{answer}\mid\text{question},\text{documents})$. Since the question, answer, and model are all fixed, the difference in answer codelength across the four conditions reflects how document-return quality affects the system's compression of the answer.

\begin{table}[t]
\centering
\caption{Answer codelength under different retrieval conditions on HotpotQA.}
\label{tab:rag}
\begin{tabular}{llrrr}
\toprule
Condition & Context & Mean codelength & Median codelength & Savings vs.\ D0 \\
\midrule
D0 & retrieval failure & $24.46$ & $22.79$ & --- \\
D1 & gold relevant docs & $6.17$ & $3.65$ & $18.29$ \\
D2 & distractor docs & $21.52$ & $19.63$ & $2.95$ \\
D3 & gold + distractor mix & $7.34$ & $5.60$ & $17.13$ \\
\bottomrule
\end{tabular}
\end{table}

\begin{figure}[htbp]
  \centering
  \begin{subfigure}[t]{0.49\textwidth}
    \centering
    \includegraphics[width=\linewidth]{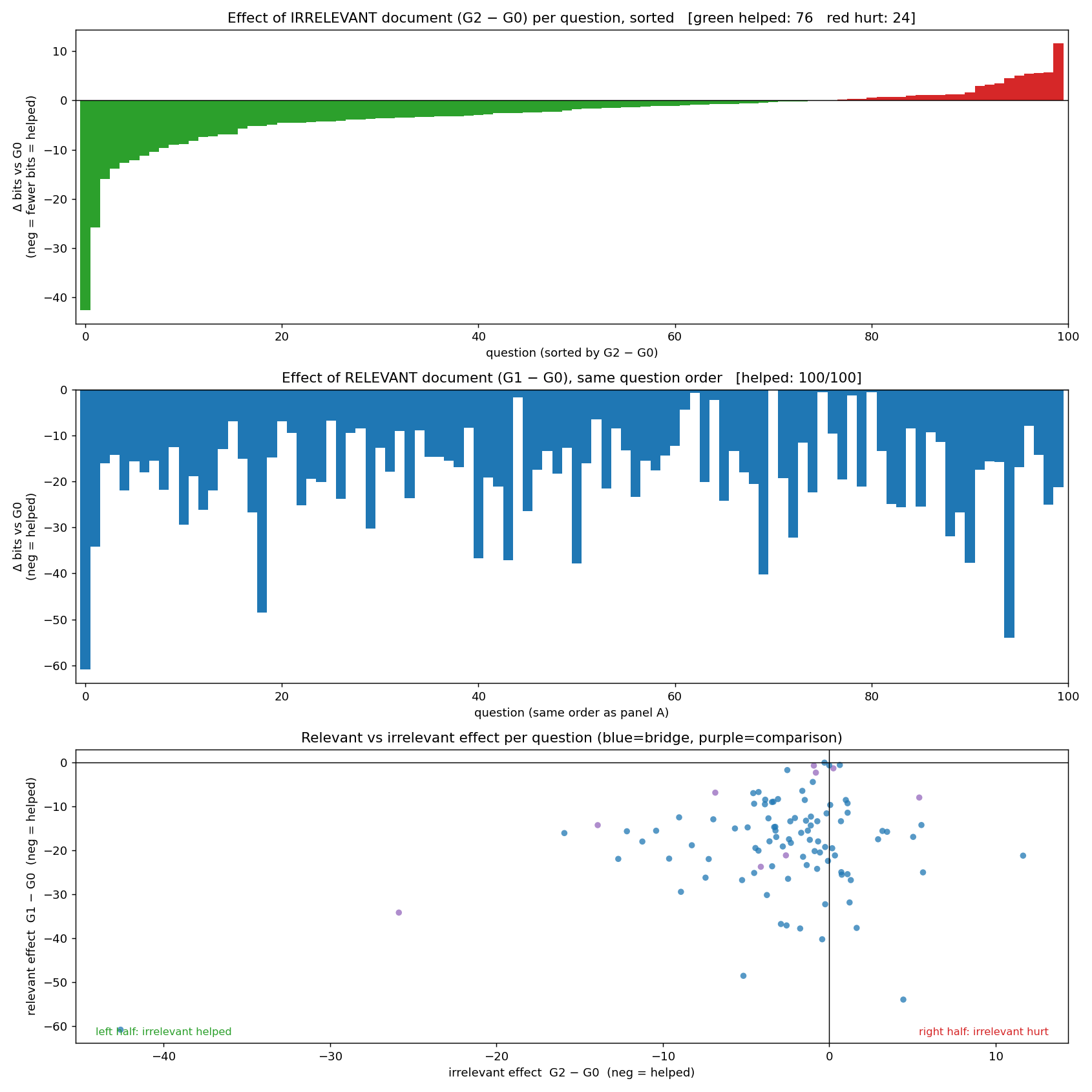}
    \caption{Per-question reference-document effect on HotpotQA. Irrelevant documents sometimes help and sometimes harm; relevant documents lower the answer codelength on all samples. The labels G0--G3 are internal markers in the experiment source, corresponding to D0--D3 in the text.}
    \label{fig:ragperq}
  \end{subfigure}
  \hfill
  \begin{subfigure}[t]{0.49\textwidth}
    \centering
    \includegraphics[width=\linewidth]{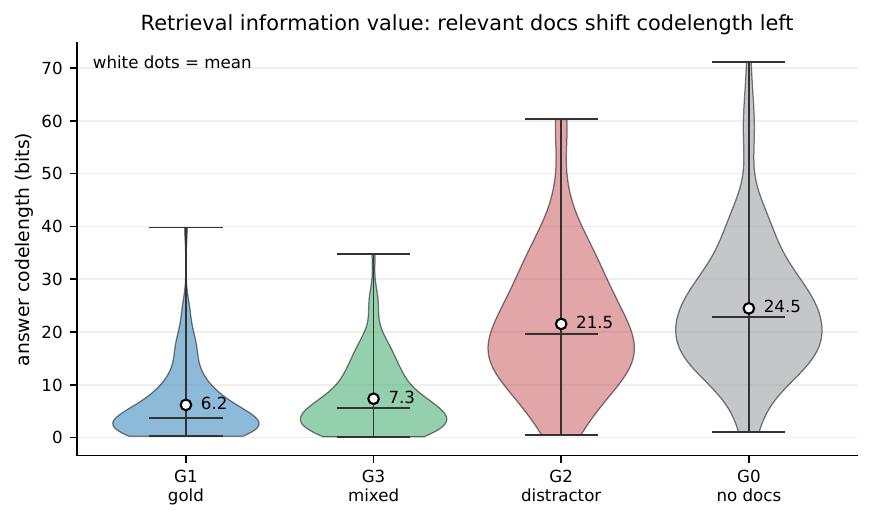}
    \caption{Distribution of the information value of retrieval. D1 and D3 shift the whole answer-codelength distribution left, while D2 still overlaps heavily with D0. White dots are means.}
    \label{fig:ragdist}
  \end{subfigure}
  \caption{Retrieval value on HotpotQA, priced in bits. (a) Per-question effect of relevant vs.\ distractor documents; (b) the resulting answer-codelength distributions.}
  \label{fig:rag}
\end{figure}

Table~\ref{tab:rag} gives the average effect, but the per-question reference-document effect better illustrates the quantitative-analysis value of the retrieval component. Figure~\ref{fig:ragperq} uses D0 as baseline to compare, per question, the codelength change of the same answer under the irrelevant document block D2 and the relevant document block D1. Irrelevant documents are not simply ``useless'': codelength drops on $76/100$ questions and rises on $24/100$, saving only $2.95$ bits on average; the maximum help is about $42.6$ bits and the maximum harm about $11.6$ bits. This shows that irrelevant passages sometimes provide a similar-topic prior and sometimes push the model toward a wrong answer. By contrast, relevant documents lower the codelength on all $100/100$ questions, saving $18.29$ bits on average; the mixed retrieval D3 also lowers it on all $100/100$ questions, showing that as long as relevant evidence is in the context, the model can usually pick out the useful information from distractor passages.

Relevant documents compress the answer codelength to about a quarter (D1, saving $18.29$ bits), distractors barely help (D2, only $2.95$ bits), and the mix retains most of the gain (D3). Figure~\ref{fig:ragdist} shows this is not a few samples dragging down the mean: the whole D1 and D3 distributions shift left, while D2 overlaps heavily with D0. Thus retrieval value is bits priced continuously by relevance, not an on/off switch.

\subsection{Claim 2: observation granularity affects compression}\label{sec:claim2}

The looser the observation standard, the more reconstructions are acceptable and the fewer bits the system needs to compress the target; the stricter it is, the more bits are needed. Thus compression ability depends on observation granularity and must be reported together with it---an isolated score cannot say to what degree the reconstruction was achieved.

\begin{figure}[htbp]
  \centering
  \includegraphics[width=0.85\linewidth]{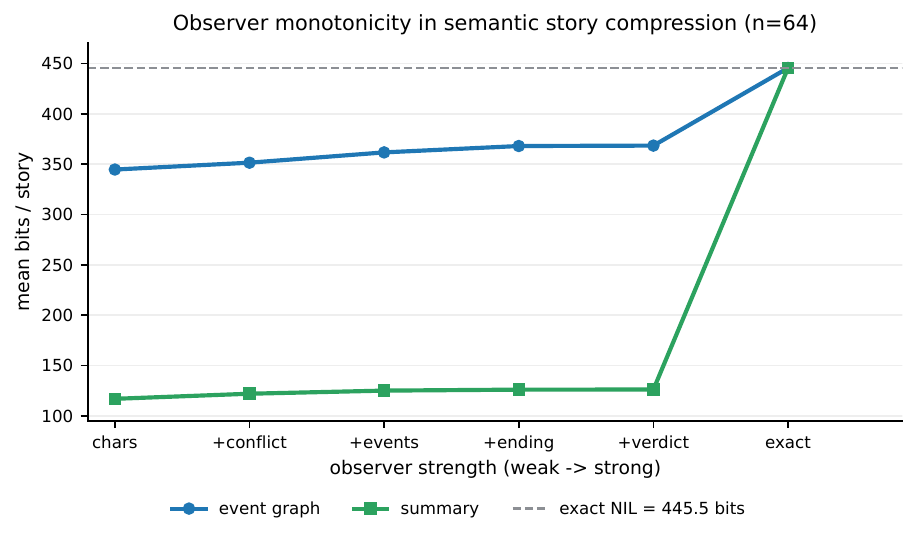}
  \caption{Observer monotonicity. The stronger the observer, the higher the mean codelength rises monotonically, approaching the exact-reconstruction ceiling.}
  \label{fig:ladder}
\end{figure}

We validate this on the story task. The system uses the \textbf{seed coding} of Section~4: it compresses a story into a short latent $z$ (a summary or event graph) plus the index of a successful seed, and the decoder replays that seed to regenerate the story from $z$, falling back to exact coding (E0) when no seed within budget yields a semantically equivalent reconstruction. As the observer ladder progressively tightens the judge's standard (from ``characters preserved'' to ``conflict, events, and ending all correct''), the acceptable set can only shrink, so the codelength should be monotonically nondecreasing. Figure~\ref{fig:ladder} confirms this: both the summary and event-graph arms rise monotonically and approach the exact-coding ceiling, with no violations---turning ``observation granularity determines compression ability'' from an assertion into a verified prediction. This experiment serves to demonstrate the semantic-observer and budget mechanisms; its conclusions are limited by the TinyStories sample size and the reliability of the LLM judge.

\subsection{Claim 3: a trade-off between compression and compute}\label{sec:claim3}

\begin{figure}[htbp]
  \centering
  \begin{subfigure}[t]{0.49\textwidth}
    \centering
    \includegraphics[width=\linewidth]{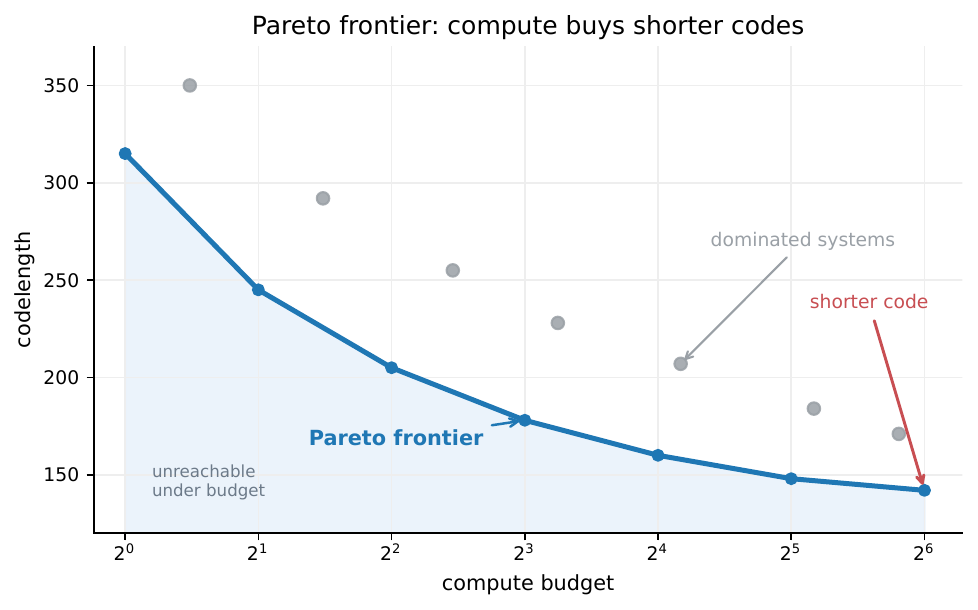}
    \caption{Schematic: a larger budget buys a shorter code, but with diminishing returns; below the frontier is unreachable, above it lie suboptimal protocols.}
    \label{fig:pareto-schem}
  \end{subfigure}
  \hfill
  \begin{subfigure}[t]{0.49\textwidth}
    \centering
    \includegraphics[width=\linewidth]{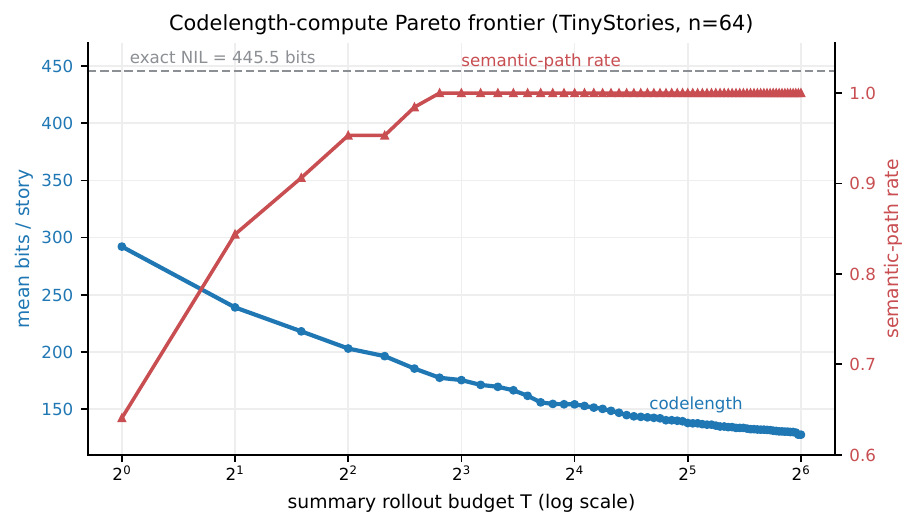}
    \caption{Measured frontier on TinyStories. The larger the rollout budget, the lower the codelength; the semantic-path rate saturates around $T=8$.}
    \label{fig:pareto}
  \end{subfigure}
  \caption{Compute--codelength Pareto frontier: (a) schematic and (b) measured on TinyStories.}
  \label{fig:pareto-both}
\end{figure}

There is a trade-off between a system's compression ability and the compute invested: an agentic system exchanges search, sampling, or verification for a shorter description. Figure~\ref{fig:pareto-schem} (schematic) gives the compute--codelength Pareto frontier---increasing the budget usually improves compression (shorter codelength) but with diminishing returns; below the frontier is unreachable, and above it lie suboptimal protocols.

\paragraph{E story (budget sweep).} Using the same seed-coding system as Section~\ref{sec:claim2}, we fix the seed budget used for reconstruction and sweep a single variable---the number $T$ of summary rollouts the encoder may search, i.e.\ how many candidate latents the encoder tries before taking the one that ``produces an accepted reconstruction with the shortest codelength itself.'' At each $T$ we measure the codelength and the \textbf{semantic-path rate} (the fraction of instances that succeed via the semantic-compression path, i.e.\ the summary path, without falling back to E0). Table~\ref{tab:budget} shows that codelength decreases monotonically as $T$ grows, but increasing the budget from $T=8$ to $T=64$ ($8\times$) only compresses the codelength from $175.3$ to $127.5$, with clear diminishing returns; the path rate saturates at $100\%$ around $T\approx8$, after which continued search mainly finds shorter latents rather than raising the acceptance rate. This curve shows that a single static capability number cannot describe how capability varies with budget.

\begin{table}[t]
\centering
\caption{Summary rollout budget and codelength on TinyStories (system E1).}
\label{tab:budget}
\footnotesize
\setlength{\tabcolsep}{4pt}
\begin{tabular}{l|rrrrrrrr|rr}
\toprule
$T$ & 1 & 2 & 4 & 6 & 8 & 16 & 32 & 64 & E0 (NIL exact) & gzip \\
\midrule
Bits/story & $292.1$ & $238.9$ & $202.9$ & $185.4$ & $175.3$ & $154.1$ & $137.8$ & $\mathbf{127.5}$ & $445.5$ & $2820.2$ \\
Semantic-path rate & $64.1\%$ & $84.4\%$ & $95.3\%$ & $98.4\%$ & $100\%$ & $100\%$ & $100\%$ & $100\%$ & --- & --- \\
\bottomrule
\end{tabular}
\end{table}

This experiment shows that a system's compression ability is not a single static number but a function curve of compute (Figure~\ref{fig:pareto}). Compared with a one-shot success rate, the codelength--compute curve simultaneously exposes the trade-offs among compression ability, cost, and engineering budget.

\section{Discussion}

\subsection{Marginal bit value vs.\ traditional ablation and success rate}
We measure a component's value by ``the improvement in system compression after providing the component''---i.e.\ the average codelength reduction. This differs fundamentally from the traditional practice of ``ablating a component and observing the change in model success rate,'' and is more informative for three reasons.

\begin{itemize}
  \item \textbf{It remains cheap to measure when the success rate is extremely low or too coarse.} Through teacher-forcing-like coding, we can read out component value even in tasks where direct sampling almost never succeeds. Protein masked reconstruction is a clear example: estimating the success rate by sampling would require an enormous number of samples, whereas a single forward pass of arithmetic coding reads out the homologous-template value of $83.85$ bits/seq. Conversely, in functionally equivalent protein generation the success rate is already measurable, but it only tells us $68\%\rightarrow98\%$; the rank codelength under shared context further shows that verifier-feedback repair removes $375.41$ bits/seq on average and localizes it to the motif-level constraint search cost.
  \item \textbf{It can distinguish regions that all look like zero from a success-rate viewpoint.} From a success-rate viewpoint, $2^{-10}$ and $2^{-100}$ are both essentially zero. From a codelength viewpoint, moving from the latter to the former gains $90$ bits of information. A component that raises the success probability from $2^{-100}$ to $2^{-10}$ looks useless on a success-rate plot but is extremely valuable in bits.
  \item \textbf{It provides a conditional, public accounting unit.} Under the same task, the same protocol, the same unit, and a fixed order of component additions, bit values can be summed to track the marginal contribution of different components to the total codelength reduction; for cross-task or cross-protocol comparisons, one must additionally report the distribution, observation standard, budget, and unit.
\end{itemize}

\subsection{The relativity of compression: intelligence cannot be discussed in isolation}

Any compression result is relative to a \textbf{distribution, observation granularity, model, environment interface, and compute constraint}. This relativity is a feature, not a nuisance, because deployed intelligence is always exercised under these conditions. Agentic codelength measures \textbf{residual information}: if the model, verifier, retriever, or environment already provides structure, then the per-instance message can be very short. This does not mean the object itself carries little information; it means that, relative to a fixed decoding protocol, the object is easy. In this sense, an agentic workflow is not merely an implementation detail but a change of description language: it moves information from the per-instance message into the public protocol and the decoding-side system. Section~\ref{sec:agentcoding} defines this residual codelength within a fixed protocol and defines component value via the average codelength difference before and after adding the component.

\subsection{Limitations}

\textbf{Protocol limitation.} Seed coding applies only to deterministic or replayable environments: the encoder and decoder must be able to reproduce the same random trajectory, tool returns, and environment state. Engineering an encoding for non-replayable external services is out of scope; but this does not affect the paper's theoretical analysis framework, because, without loss of generality, we can in principle abstract the randomness of any environment and uncontrollable external inputs as a pseudorandom sequence pre-generated from a sufficiently long global seed. This idealized replay assumption lets the paper's analysis framework apply, without modification, to systems in non-replayable environments.

\textbf{Experimental limitation.} The five sets of experiments cover several component types typical of real agentic systems and demonstrate the ability of codelength to analyze component value, observation granularity, and the budget trade-off; but the experiments are small in scale and mainly serve to verify the paper's claims and provide exploratory evidence. These results show that the framework can analyze the mechanisms of real systems and offer methodological inspiration for analyzing larger-scale agentic systems; but this paper has not yet conducted a full, large-scale end-to-end system experiment, and the relevant numbers should not be directly extrapolated to general conclusions for deployed systems.

\subsection{Agentic pseudo-entropy}

The preceding text mainly discusses the average codelength under a fixed protocol; but if our concern is not some already-realized system, but rather ``given a model, environment, and resource constraints, how much can any designable agentic program at best compress the distribution,'' then we must generalize the measure from a single protocol to a class of constrained systems. Fixing the task distribution $\mathcal{D}$, the model $M$, and the set of environment interfaces $E$, and allowing only the program $P$ to vary, we obtain a class $\mathfrak{A}$ of constrained agentic systems satisfying constraint $C$. On top of the codelength in Section~\ref{sec:agentcoding}, taking the infimum over the system class and the expectation over the distribution, we can define the \textbf{agentic pseudo-entropy}
\[
H^{\mathfrak{A},C}_{\mathrm{agentic}}(\mathcal{D})
=
\inf_{\mathcal{A}\in\mathfrak{A}}\
\mathbb{E}_{(c,x)\sim\mathcal{D}}\,
L^{C}_{\mathcal{A}}(x\mid c).
\]
This definition follows the idea of Yao-type pseudo-entropy \citep{yao1982trapdoor,barak2003computational}: complexity is not the absolute Shannon entropy of the object distribution, but incompressibility relative to a class of constrained systems. Its main significance is to give a \textbf{system-relative data complexity}: as seen by the constrained system class induced by a fixed model, environment interface, and resource constraints, how much true complexity of the distribution $\mathcal{D}$ remains that cannot be absorbed by any program, tool, retrieval, search, or verifier. In other words, agentic pseudo-entropy asks not ``how much information'' the data themselves carry, but how hard the data still are to compress within the capability boundary of this class of agentic systems.

From a system perspective, the same quantity also characterizes the optimal achievable compression ability of the system class on the distribution $\mathcal{D}$ given the model, environment interface, and resource constraints---i.e.\ the theoretical upper bound on learning or exploiting the structure of the distribution. If the agentic pseudo-entropy of some distribution remains high, it means that under these fixed resources the distribution is indeed hard for this system class; if it is low, much of the distribution's structure can already be exploited by the existing model, environment interface, and admissible programs.

The constraint $C$ here must include limits such as program length and compute, e.g.\ program length, number of model calls, number of tool calls, or call cost (an engineerable agentic system naturally satisfies these constraints). These constraints are constitutive: without limiting program length, the system could write the complex structure that should be borne by the model directly into the program, even explicitly unrolling the model's internal computation, making the definition degenerate. Without limiting compute, some tasks could shift description length into unbounded search; as Appendix~\ref{app:resource} shows, in certain public-specification tasks the decoder can find any qualifying artifact by infinite enumeration, making the per-instance codelength degenerate.

This paper has not yet fully explored the definition of agentic pseudo-entropy itself or the details of these constraint settings; we present it here only as a heuristic concept, to suggest that data complexity relative to a concrete model/system can be discussed from the perspective of a constrained system class.

\section{Conclusion}

We introduced a compression-based method for analyzing system capability at the level of agentic systems. The key step is to fix the task distribution, observer, interface, and budget, and then ask: how many residual bits must each instance still supply for reconstruction. This makes tools, environment, retrieval, verifiers, and the search budget part of the system being analyzed, rather than external implementation details.

The viewpoint becomes operational through \textbf{arithmetic coding}, \textbf{seed coding} in deterministic or replayable environments, and \textbf{fallback}. Each component is measured by the residual bits it removes---its average codelength reduction (values in Table~\ref{tab:claims}). The results show that treating an agentic system as a shared decoder can explain and quantify part of where system capability comes from. This is not the only definition of intelligence, but a complementary viewpoint for component analysis, observer analysis, and budget analysis; it suggests which shared structures may reduce the residual description length, while the concrete engineering significance still needs to be assessed relative to a specific system.

\section*{Acknowledgments}
This work was supported by the National Natural Science Foundation of China (Grant No.\ 625B1012).

\bibliography{references}

\appendix
\section{Experimental Implementation Details}\label{app:exp}

Below we describe the full design of the five experiments mentioned in Section~5.

\subsection{Reversed text (Section~\ref{sec:claim1})}\label{app:reverse}

\textbf{Data, model, and sampling scale.} The target distribution is the first $1\,\mathrm{MiB}$ ($1{,}048{,}576$ bytes) of natural text from enwik8 (the English Wikipedia dump), with a \textbf{deterministic line-wise character reversal} $R$ applied: characters within a line are reversed as a whole while line terminators are preserved (e.g.\ \texttt{"Hello World\textbackslash n"} $\mapsto$ \texttt{"dlroW olleH\textbackslash n"}). $R$ is an involution (applying it twice restores the original). The model is the SmolLM2-360M base version. The observation family $\mathcal{O}$ requires exact byte equality (lossless recovery). This task is a single-sample whole-segment encoding, with no sampling and no fallback.

\textbf{Agentic system construction.} The only ``tool'' the system has is the deterministic reversal operator $R$, shared by the encoder and decoder. A0 does not use it and encodes the reversed string directly; A1 uses it, hardwiring the per-instance information ``this text is reversed'' into the shared transform so it need not be carried in the bitstream.

\textbf{Codelength protocol.} The two systems share the same model and (in A1) the reverse tool, differing only in where the information is placed.

\textit{A0 (no tool, encode reversed string directly).} Encoder: arithmetic-code the \textbf{reversed} target string $x$ directly. Decoder: arithmetic-decode to obtain $x$. Stored code: just one arithmetic code of $x$; no tool, no seed, no fallback.

\textit{A1 (reverse tool).} Encoder: first use the shared tool to compute the canonical text $u=R(x)$, then arithmetic-code $u$, $L_{\mathrm{A1}}(x)=L_{\mathrm{LM}}(u)+O(1)\ll L_{\mathrm{A0}}(x)$. Decoder: first arithmetic-decode $u$, then apply the same shared tool to output $x=R(u)$---the reversal step is performed by the decoder and so does not occupy the bitstream. Stored code: just one arithmetic code of $u$; the reversal rule $R$ and the model weights are shared fixed components and are not stored.

\subsection{Chess (Section~\ref{sec:claim1})}\label{app:chess}

\textbf{Data, model, and sampling scale.} The target distribution is the January 2013 shard of the Lichess standard game database, deterministically sampled with a fixed seed and keeping only games whose half-move count lies in $[20,120]$, giving $100$ games and $6{,}005$ half-moves in total. Moves are in standard algebraic notation (SAN). The model is SmolLM2-360M-Instruct. The observation family $\mathcal{O}$ requires exact trajectory agreement.

\textbf{Agentic system construction.} The system's tool is a \textbf{legal-move environment} (\texttt{python-chess}): at any position it enumerates the full set $L_t$ of legal SAN moves. B0 does not use this tool; B1 uses it at each step to renormalize the model distribution over the legal set, recovering the probability mass that would otherwise leak to illegal moves. The rule engine and the initial board are equally visible to both ends, and the environment calls are shared computation.

\textbf{Codelength protocol.}

\textit{B0 (no tool, encode the whole game directly).} Encoder: concatenate the whole game's SAN into a space-separated text string and arithmetic-code it token-by-token. Decoder: arithmetic-decode the whole SAN text stream token-by-token. Stored code: one arithmetic code of the whole SAN string; no environment call, no fallback.

\textit{B1 (legal-move environment).} Encoder: at each position $t$, call the environment to get the legal set $L_t$, score each candidate move's continuation $s_a$, renormalize over the legal set by log-sum-exp, and take the observed move $a^\star$'s codelength as $\operatorname{logsumexp}_a(s_a)-s_{a^\star}=-\log_2 Q(a^\star\mid h_t,L_t)$, where $Q(a\mid h_t,L_t)=P_M(a\mid h_t)/P_M(L_t\mid h_t)$ (since $L_t$ is a legal subset, $P_M(L_t\mid h_t)\le1$, so each step's codelength does not increase relative to the bare model, i.e.\ $\Delta_{\mathrm{legal}}\ge0$). Decoder: replay the whole game step-by-step; at position $t$, use the same history $h_t$ to call the same engine for the same $L_t$, recompute the same renormalized distribution $Q$, arithmetic-decode the step's move from the bitstream, and advance to $h_{t+1}$. Stored code: one step-by-step arithmetic bitstream, which is read step-by-step against the legal sets replayed by the decoder to restore the whole trajectory; the rule engine, notation, and weights are shared fixed components and are not stored.

\subsection{Protein (Section~\ref{sec:claim1}, masked ESM)}\label{app:protein}

\textbf{Data, model, and sampling scale.} The target distribution is UniProt sequences with Pfam annotations and length $80\!-\!300\,\mathrm{aa}$; the core experiment takes $50$ each from PF00069 (protein kinase domain) and PF00096 (C2H2 zinc finger), $100$ in total. The model is the masked language model ESM2 (\texttt{esm2\_t33\_650M\_UR50D}). To eliminate leakage, we use \textbf{leave-one-out}: when processing target sequence $x$, the reference set used for the template library, motif extraction, and verifier checks is always ``same-family sequences minus $x$ itself.'' The observation family $\mathcal{O}$ is \textbf{functional equivalence}: a fixed lightweight verifier accepts only when all three hold---length in $[80,300]$, the candidate contains the top-3 3-mer motifs counted from the reference set, and a 3-mer Jaccard similarity above $0.2$ with some reference---rather than requiring residue-by-residue identity. The actual codelength experiment uses a masked conditional reconstruction proxy, billing only the masked residues.

\textbf{Agentic system construction.} Beyond ESM, we add a \textbf{homologous-template tool}: it retrieves and locally aligns a homologous template from the leave-one-out reference set, obtaining a per-position template prior $p_{\text{prior}}$, and at each masked position linearly mixes it with ESM's own distribution as $(1-\alpha)\,p_{\text{ESM}}+\alpha\,p_{\text{prior}}$. C0 does not use this tool; the tool has two versions: \textbf{C1 fixed} assumes all templates are equally reliable (constant match probability, $\alpha$ capped at $0.50$); \textbf{C2 calibrated} estimates each template's per-position match reliability \textbf{from visible (unmasked) sites only}, and accordingly relaxes the mixing cap to $\alpha\le0.95$---the more reliable the template, the more it is mixed in. The mixing coefficient $\alpha$ is selected on held-out visible residues.

\textbf{Codelength protocol.} The three systems share the same set of mask positions (selected at a fixed $30\%$ ratio, protecting motif sites), and the codelength is the masked pseudo-NLL on masked positions (mask each position and sum the $-\log_2 P$ of the corresponding distribution); visible sites are always conditions and not billed, so the difference between systems is a clean causal comparison. The three differ only in the distribution $Q$ used at masked positions. The exact length $|x|\log_2 20$ serves only as the denominator of the compression rate and does not enter the component value. This protocol estimates the template's reduction of conditional residue uncertainty; it is not a full de novo protein generation protocol.

\textit{C0 (plain ESM, no-tool baseline).} Encoder: arithmetic-code masked positions with ESM's own distribution $p_{\text{ESM}}$. Decoder: run the same ESM forward pass conditioned on the same visible sites and mask set, reproduce $p_{\text{ESM}}$, and arithmetic-decode position by position. Stored code: the arithmetic code of the masked residues under $p_{\text{ESM}}$.

\textit{C1 (fixed homologous-template tool).} Encoder: arithmetic-code each masked position with the mixture $(1-\alpha)\,p_{\text{ESM}}+\alpha\,p_{\text{prior}}$, with fixed template reliability and $\alpha\le0.50$. Decoder: re-run the homologous-template tool on the same leave-one-out reference set, reproduce the same mixture, and decode position by position. Stored code: the arithmetic code of the masked residues under this mixture; the reference set and $\alpha$ are shared conditions and not stored.

\textit{C2 (calibrated homologous-template tool).} Same encode/decode skeleton as the fixed version, except the per-position $\alpha$ of the mixture is determined by the template reliability estimated \textbf{from visible sites}, with the cap relaxed to $0.95$ (the more reliable the template, the more it is mixed in). Crucially, both ends recompute $\alpha$ from visible sites with the same rule, so no transmission is needed. Stored code: the arithmetic code of the masked residues under the calibrated mixture.

\subsection{Retrieval augmentation (Section~\ref{sec:claim1}, HotpotQA)}\label{app:rag}

\textbf{Data, model, and sampling scale.} Samples are taken from HotpotQA's distractor configuration, validation split, removing yes/no answers and requiring both gold (supporting) and distractor passages, taking the first $100$ (bridge $91$, comparison $9$); each item has $2$ gold and $8$ distractor passages. The model is Qwen3.5-0.8B. The observation family $\mathcal{O}$ is the exact answer text, held consistent across the four conditions of the same sample. Here the \textbf{public condition $c$ is the question itself} (the problem statement), and the answer is conditionally compressed given the question; the reference documents are \textbf{not} $c$ but the output of the \textbf{retrieval component} being measured---a decoding resource determined by the question and shared by both ends (in the same class as the model weights and the rule engine). The four conditions D0--D3 are controlled oracle document-return conditions and thus constitute four \textbf{agentic systems differing only in the retrieval component's returned content}, isomorphic to A0/A1 and B0/B1.

\textbf{Agentic system construction.} The system has a controlled \textbf{retriever} whose output (reference documents) is injected into the model context to provide information. The four conditions differ only in the relevance of the documents the retriever returns: D0 a retrieval-failure placeholder (no valid documents); D1 only the $2$ gold (supporting) passages; D2 $2$ distractor passages equal in number to gold; D3 all $10$ passages in original order. This setup measures how document-return quality affects answer codelength.

\textbf{Codelength protocol.} D0--D3 are four instances of the same class of retrieval-augmented system, differing only in the retrieval component; the encoder, decoder, and stored-code skeleton are identical, so the protocol is described once.
\begin{itemize}
  \item \textit{Encoder design.} The retriever's output reference documents are placed in the prefix, and the conditional NLL $-\log_2 P(\text{answer}\mid\text{question},\text{documents})$ is accumulated and arithmetic-coded \textbf{over the answer tokens only}; the prompt (including the question and reference passages) is a shared prefix and is not billed at all, forming a clean causal comparison across conditions.
  \item \textit{Decoder design.} The decoder obtains the same documents through the same retriever (the same question), constructs the same prefix, runs the same forward pass, and arithmetic-decodes the answer token-by-token---just as the chess decoder in \ref{app:chess} recomputes the legal set itself.
  \item \textit{Stored code.} Just one arithmetic code of the answer; the reference documents, like the model weights, are shared components and not transmitted per instance, hence not stored---the retrieval component's value is precisely how many bits it shortens this answer code by.
  \item \textit{Per-question reference-document effect analysis.} \texttt{analyze\_per\_question.py} reads each question's four-condition answer codelengths from \texttt{per\_sample\_metrics.csv} and computes $\Delta_{\mathrm{irrel}}=L_{\mathrm{D2}}-L_{\mathrm{D0}}$ and $\Delta_{\mathrm{rel}}=L_{\mathrm{D1}}-L_{\mathrm{D0}}$; a negative value means the document block lowers the codelength, a positive value means it harms compression. Figure~\ref{fig:ragperq} in the main text is plotted sorted by $\Delta_{\mathrm{irrel}}$, showing the relevant-document effect under the same question order.
\end{itemize}

\subsection{Story (Sections~\ref{sec:claim2} and \ref{sec:claim3}, TinyStories)}\label{app:story}

\textbf{Data, model, and sampling scale.} The target distribution is TinyStories, first filtered to short pieces of $55\!-\!160$ words, shuffled with a fixed seed, and taking $64$ pieces. The generator is Qwen3.5-0.8B. The observation family $\mathcal{O}$ is \textbf{semantic plot equivalence} (whether the main characters, main events, conflict, and ending are all preserved), judged via API by a DeepSeek judge (\texttt{deepseek-v4-flash}). The judge returns structured flags characters\_preserved, conflict\_preserved, main\_events\_preserved, ending\_preserved, and an overall verdict; \texttt{verdict} equal to \texttt{correct} is judged as the strongest semantic acceptance.

\textbf{Observer ladder.} To test observation-granularity monotonicity, we construct a weak-to-strong cumulative-AND chain from the judge's flags: characters preserved; characters + conflict preserved; characters + conflict + main events preserved; characters + conflict + main events + ending preserved; plus the overall \texttt{verdict=correct}. Each rung only adds constraints, so a stronger observer's acceptable set is contained in that of a weaker observer. This ladder is recomputed offline entirely from the saved per-seed judge results; the DeepSeek judge compares the original story with the candidate reconstruction each time (we acknowledge that the noise of this judge model may have an effect, which is one limitation of our experiments).

\textbf{Agentic system construction.} The encoder compresses a whole story into a short \textbf{latent descriptor}, and the decoder \textbf{regenerates} the story from it; seed coding lets the intermediate reconstruction avoid token-by-token coding, and the NIL fallback (baseline E0) guarantees the worst case is no worse than bare coding. Two billed latents: E1 is a \textbf{free-form summary}, and E2 is a \textbf{structured event graph} (flat JSON, incorporated into the decoding protocol via a fixed parser/normalizer). Both latents are billed by arithmetic coding with the \textbf{empty context (NIL)}, i.e.\ $-\log_2 P(\text{latent}\mid\text{NIL prompt})$.

\textbf{Codelength protocol (seed coding + fallback).} E1 and E2 share the same seed-coding + fallback skeleton, differing only in latent type (E1 free-form summary; E2 structured event-graph JSON, with one extra fixed parse/normalize step on the decoding end), so the skeleton is described once with the two differences noted.
\begin{itemize}
  \item \textit{Encoder design.} First arithmetic-code the latent $z$ by empty-context NLL (E1's $z$ is the summary, E2's $z$ is the event-graph JSON). Then enumerate $8$ public seeds (ranks $0\!-\!7$): each seed deterministically regenerates a candidate story, which is verified by the judge as $o(\hat x)=o(x)$, and take the rank of the first accepted seed; for E1, additionally select, among multiple accepted summary candidates, the one with the shortest empty-context NLL. If none of the $8$ seeds is accepted $\rightarrow$ fall back to \textbf{NIL (E0)}: bill by the empty-context arithmetic code of the whole story.
  \item \textit{Decoder design.} First read the $1$-bit method selector. Semantic path: arithmetic-decode the latent $z$ (E2 normalized to a canonical latent by the fixed parser), read the $3$-bit seed rank, and deterministically regenerate the story with that seed (acceptance was verified at encoding time, and the decoding end reproduces the same reconstruction position-by-position). NIL path: directly arithmetic-decode the whole story.
  \item \textit{Stored code.} A $1$-bit method selector, plus one of: the semantic path stores [latent empty-context code $+$ seed rank code]; the fallback path stores [whole-story NIL code]. Under the budget-sweep convention of Section~\ref{sec:claim3}, the reconstruction seed budget is $8$ (ranks $0\!-\!7$), and the rank uses a fixed $3$-bit code; if the selector is counted on both paths, the combined code exceeds E0 (NIL) by at most $1$ bit, and is below E0 when the semantic path succeeds and is shorter. The seed set, generator weights, and (E2's) parser are shared fixed components and not stored. The compute--codelength Pareto frontier is obtained under this protocol by sweeping the summary rollout budget $T=1\dots64$ (replaying only the first $T$ rollouts) (values in Section~\ref{sec:claim3}).
\end{itemize}

\section{Technical Notes and Proof Sketches}\label{app:tech}

This appendix collects the short technical facts used in the main text. Throughout, $L$ denotes the operational codelength under a fixed protocol.

\subsection{Arithmetic coding and the NLL proxy differ by a per-instance constant}\label{app:ac}
Under context $c$ (and side information $z$), the model induces conditional distributions $P_M(\cdot\mid x_{<t},c)$ in token order. Arithmetic coding maps the reference $x=x_{1:n}$ to a \textbf{nested} subinterval of the unit interval: starting from $[0,1)$, after reading $x_t$ it subdivides the current interval proportionally according to the current token distribution and takes the segment corresponding to $x_t$; after $n$ steps it obtains
\[
I_x=\big[F(x),\,F(x)+P_{\mathcal{A}}(x\mid c,z)\big),\qquad |I_x|=P_{\mathcal{A}}(x\mid c,z)=\prod_{t}P_M(x_t\mid x_{<t},c),
\]
where $F$ is the cumulative distribution in token order. Write
\[
\mathrm{NLL}(x)=-\log_2 P_{\mathcal{A}}(x\mid c,z).
\]
To obtain a prefix-unambiguous finite binary codeword, take the interval midpoint
\[
m_x=F(x)+\tfrac12 P_{\mathcal{A}}(x\mid c,z),
\]
and send the first
\[
\ell(x)=\left\lceil \log_2\frac{1}{P_{\mathcal{A}}(x\mid c,z)}\right\rceil+1
=\lceil\mathrm{NLL}(x)\rceil+1
\]
binary digits of $m_x$. This is exactly the standard length choice in Shannon--Fano--Elias codes and in the termination analysis of arithmetic coding: rounding up ensures the binary codeword is fine enough, and the extra $+1$ makes the small binary interval determined by this prefix lie entirely within $I_x$, preserving prefix-unambiguity; \citet{deletang2023language} use the same convention when compressing sequences with a predictive distribution $\rho$, writing the codelength as $\lceil-\log_2\rho(x_{1:n})\rceil+1$. The decoder shares $c,z$ and the same model, recomputes the same nested subdivision for the known length $n$, and can thus uniquely locate $I_x$ from this prefix and recover $x$. Hence
\[
1\ \le\ \ell(x)-\mathrm{NLL}(x)\ <\ 2,
\]
where the left end is attained when $\mathrm{NLL}(x)$ is an integer; if only an upper bound is of interest, then $\ell(x)\le \mathrm{NLL}(x)+2$. Therefore, when we use NLL as a codelength proxy, we ignore the prefix-termination constant of fewer than $2$ bits per message; all reported component values and average codelength differences are far larger than this per-instance constant, and the conclusions are unaffected \citep{cover2006elements,witten1987arithmetic,deletang2023language}. $\square$

\subsection{Resource constraints are a constitutive part of the definition}\label{app:resource}
Agentic codelength measures how many extra bits the encoder must still send for this instance after the public condition $c$, the shared system components, and the observation standard have all been fixed. This definition is meaningful only when compute is also limited; otherwise, the decoder could offload ``finding a qualifying artifact among many candidates'' to unbounded enumeration, rather than having the encoder point to that artifact with bits.

This is especially crucial in \textbf{public-specification tasks}. Many agentic tasks do not require recovering a unique target $x$, but rather, given a problem, tests, rules, or specification $c$, output any $y$ satisfying the specification. For example, a QA task needs only a correct answer, a coding task needs only to pass the tests, and a math problem needs only a correct solution. Here the success judgment is not ``restore the properties of this specific $x$'' but an absolute judgment depending only on the public specification $c$,
\[
V_c(y)=1,
\]
where the original target $x$ is merely one example satisfying the specification, not an object that must be matched item-by-item. Thus, for the decoder, the task is not ``recover $x$'' but ``output any $y$ satisfying $V_c$.''

Suppose the decoder, generator, verifier, and seed-enumeration order are all shared fixed resources. If there exists a seed $r$ such that $\mathcal{A}(c,r)$ passes verification, i.e.\ $V_c(\mathcal{A}(c,r))=1$, then under unbounded compute the decoder need not receive any per-instance message; it only enumerates public seeds
\[
r=1,2,\dots
\]
and stops at the first output satisfying $V_c(\mathcal{A}(c,r))=1$. Since the acceptance criterion only requires a ``qualifying artifact'' and not a match to a specific target $x$, the first qualifying artifact the decoder outputs is already an acceptable reconstruction. Therefore such tasks can exhibit \textbf{zero-bit reconstruction} under unbounded search.

\end{document}